\documentclass[letterpaper, 10 pt, conference]{ieeeconf}  
\IEEEoverridecommandlockouts                              
\usepackage{graphics} 
\usepackage{epsfig} 
\usepackage{mathptmx} 
\usepackage{times} 
\usepackage{amsmath} 
\usepackage{amsfonts} 
\usepackage{amssymb}  
\usepackage{float}
\usepackage{bm}
\usepackage{tabularx}
\usepackage{booktabs}
\usepackage{tikzpagenodes}
\usepackage[hidelinks]{hyperref}

\usepackage{microtype}
\makeatletter
\MT@addto@setup{%
	\DeclareRobustCommand\microtypecontext[1]{%
		\MT@setup@contexts
		\let\MT@reset@context\relax
		\let\glb@currsize\@empty 
		\setkeys{MTC}{#1}%
		\selectfont
		\MT@reset@context
	}%
}
\makeatother

\usepackage[moderate]{savetrees}

\title{\LARGE \bf Stable Electromyographic Sequence Prediction During Movement Transitions using Temporal Convolutional Networks} 

\author{Joseph L. Betthauser, John T. Krall, Rahul R. Kaliki, Matthew S. Fifer, and Nitish V. Thakor
	\thanks{This work was supported by the Johns Hopkins University Applied Physics Laboratory Graduate Research Fellowship}
	\thanks{$^{}$J. Betthauser and Dr. Thakor are with the Department of Electrical and Computer Engineering, The Johns Hopkins University, Baltimore, MD 21218, USA, {\small jbettha1@jhu.edu}}%
	\thanks{$^{}$J. Krall and Dr. Thakor are with the Department of Biomedical Engineering, The Johns Hopkins University, Baltimore, MD 21218, USA}%
	\thanks{$^{}$Dr. R. Kaliki is with Infinite Biomedical Technologies, LLC., Baltimore, MD 21218, USA}
	\thanks{$^{}$Dr. Fifer is with the Research and Exploratory Development Department, Johns Hopkins University Applied Physics Lab, Laurel, MD 20723, USA}%
} 

\begin{document}
	\maketitle	
	  	\begin{tikzpicture}[remember picture,overlay]		
			\node[align=center,text=black] at ([yshift=-2.7em]current page text area.south) {\footnotesize Copyright (c) 2019 IEEE. Personal use of this material is permitted. Permission from IEEE must be obtained for all other uses,\\\footnotesize in any current or future media, including reprinting/republishing this material for advertising or promotional purposes, creating new\\\footnotesize collective works, for resale or redistribution to servers or lists, or reuse of any copyrighted component of this work in other works.};
		\end{tikzpicture}%
	\thispagestyle{empty}
	\pagestyle{empty}
	
\begin{abstract}
	Transient muscle movements influence the tem-poral structure of myoelectric signal patterns, often leading to unstable prediction behavior from movement-pattern classification methods. We show that temporal convolutional network sequential models leverage the myoelectric signal's history to discover contextual temporal features that aid in correctly predicting movement intentions, especially during interclass transitions. We demonstrate myoelectric classification using temporal convolutional networks to effect 3 simultaneous hand and wrist degrees-of-freedom in an experiment involving nine human-subjects. Temporal convolutional networks yield significant $(p<0.001)$ performance improvements over other state-of-the-art methods in terms of both classification accuracy and stability.
\end{abstract}	

\section{Introduction}
	Classification of electromyographic (EMG) signals for upper-limb prosthesis activation involves algorithmically learning to distinguish patterns in EMG signals that correlate to discrete hand, wrist, or arm motions~\cite{Hudgins93}. Techniques such as support vector machines (SVM) can discriminate a large number of EMG movement-class patterns under ideal conditions~\cite{Atzori16}. EMG pattern classification algorithms typically use individual data samples, represented as extracted features from a fixed window of raw EMG, to compute the boundaries that best segment the samples into distinct movement classes. 
	\begin{figure}[t!]
		\centering
		\includegraphics[width=0.467\textwidth]{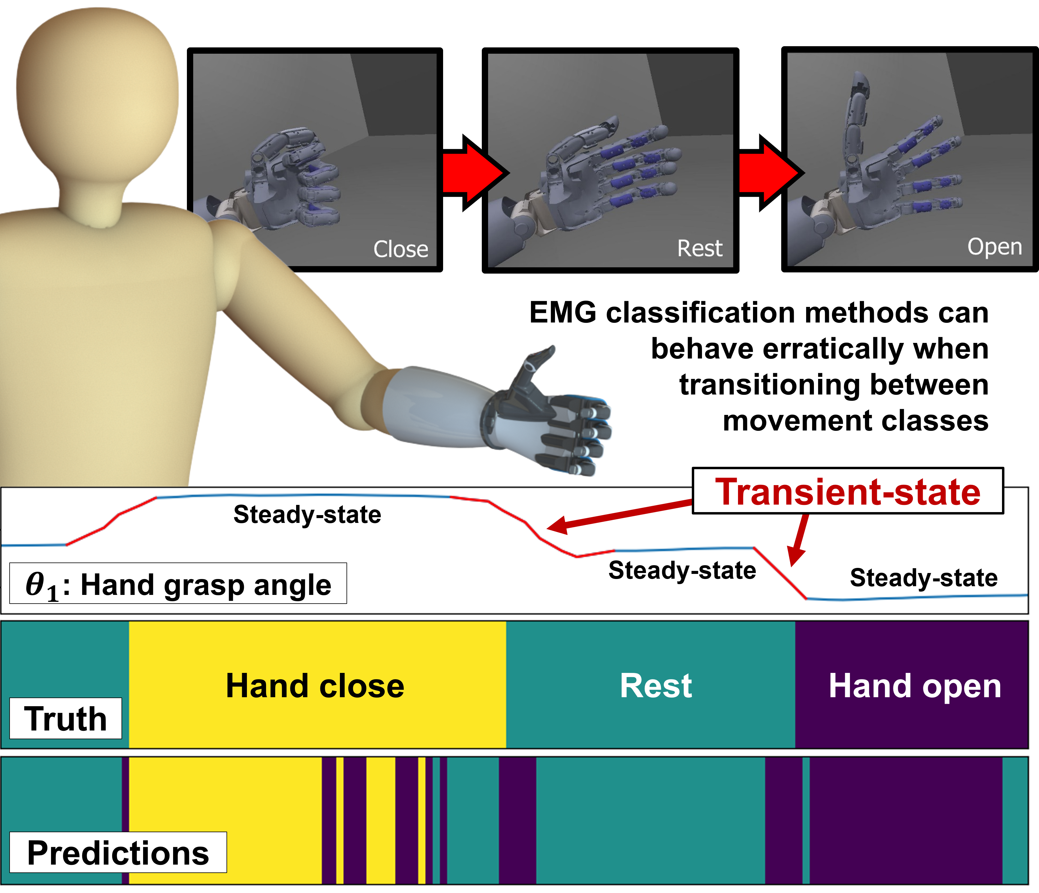}
		\caption{EMG movement-pattern classification strategies can exhibit erratic prediction behavior during transient-states when a subject is switching between classes. Steady-states during class contractions tend to elicit a more reliable, stable classifier response; however, this behavior is not guaranteed and is largely based on the subject's experience level. We propose temporal convolutional networks to improve both accuracy and stability.}
		\label{problems}    
	\end{figure}
		
	Muscle activations during \emph{steady-state} contractions are generally more \emph{stable}, especially with user practice, and the EMG signal patterns tend to reliably fall into established classes~\cite{Yang12}. However, \emph{transient-state} interclass movements pose a challenge for classifiers due to the non-stationary nature of the signal patterns~\cite{Kanitz18}\cite{Lorrain11}. In these cases, the model's prediction stream can exhibit segments of erratic and incorrect predictions (Fig.~\ref{problems}). Post-processing methods have been proposed to stabilize the prediction stream such as majority filtering and confidence-based rejection~\cite{Scheme13}. Other methods achieve stable and accurate performance by updating class boundaries adaptively~\cite{Amsuss14}\cite{Zhu17} and enhancing condition-tolerance~\cite{Betthauser18}. 
	
	Non-sequential prediction models like SVM can behave erratically during transient-states, in part because they are denied temporal context provided by the preceding sequence of consecutive inputs. EMG windows or \emph{frames} are typically predicted independently, with the windowed feature extraction technique itself serving as a compressed temporal representation of EMG. Much like photographs only capture a portion of the information about a moving subject, these models provide a rough snapshot of a dynamic system frozen in time-- critical temporal context is lost in the translation.
	
	Sequential prediction models such as long short-term memory (LSTM) recurrent networks~\cite{Hochreiter97} are the state-of-the-art for time-series prediction tasks like speech~\cite{Graves05} and activity~\cite{Javier16} recognition. At present, recurrent networks are being used for movement prediction from cortical signals~\cite{Gallego17} and EMG~\cite{Xia18}.	Herein, we present a temporal convolutional network sequential model~\cite{Lea16} for EMG classification that is significantly more accurate and stable than prevailing sequential and non-sequential methods, especially during movement transitions. 
		

\section{Methods}
\begin{figure}[t!]
	\centering
	\includegraphics[width=0.48\textwidth]{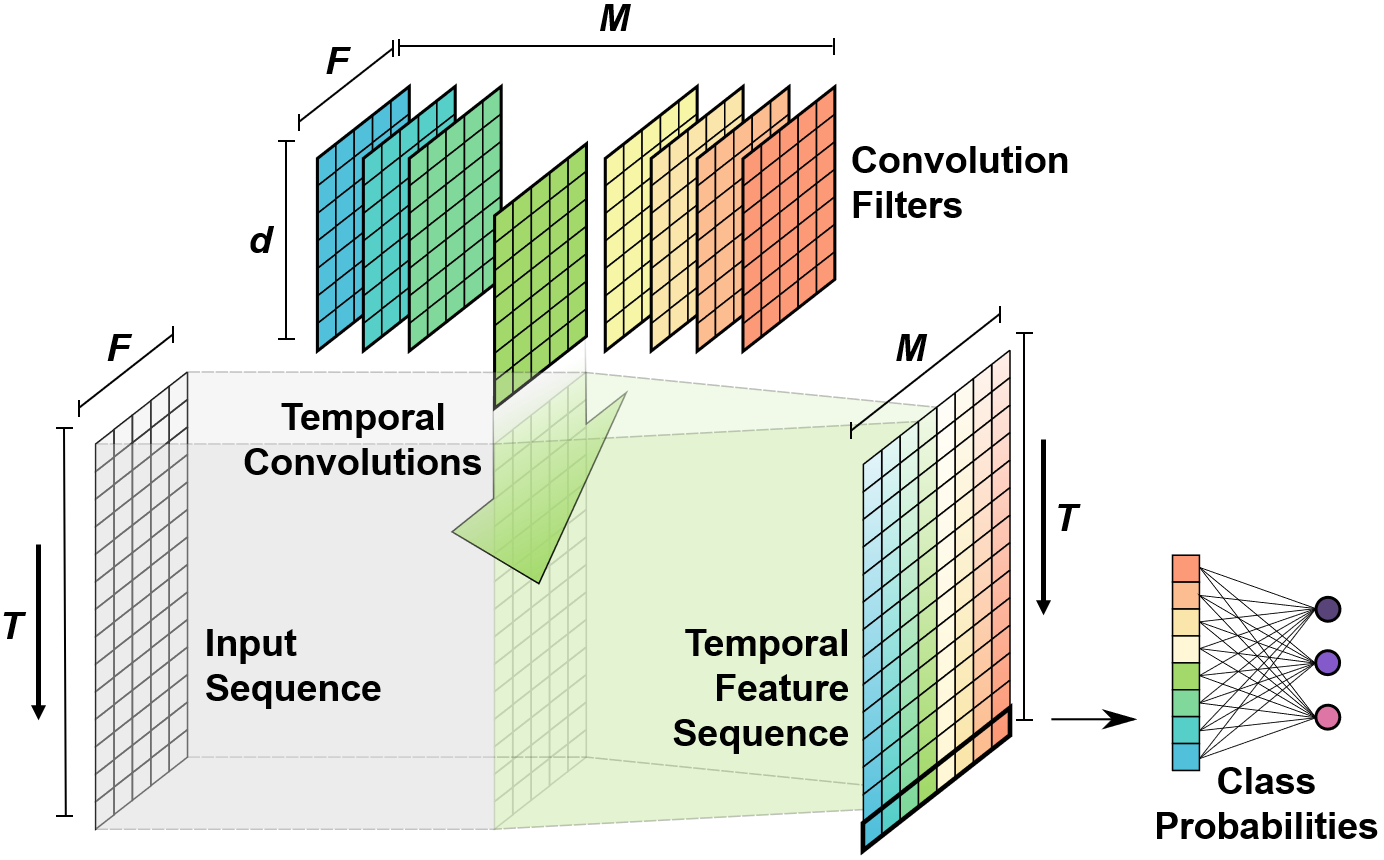}
	\caption{Multi-channel EMG features can be fed as fixed-length sequences into a TCN network for classification. TCN achieves good performance by learning convolutions which exploit temporal dependencies within the input sequences. These convolutions may provide a higher degree of regularization than other sequential models, such as the accurate but less-stable LSTM.}
	\label{model}    
\end{figure}
\subsection{Temporal Convolutional Networks}	
\begin{figure}[t!]
		\centering
		\includegraphics[width=0.48\textwidth]{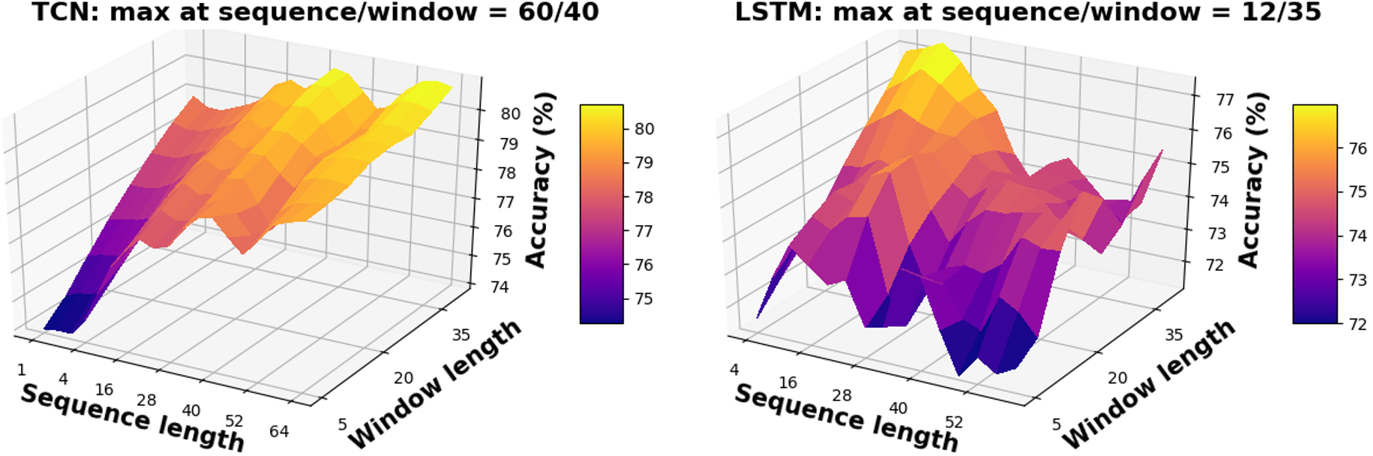}
		\caption{Performance contours for parametric tuning of TCN to determine optimal sliding window size and sequence length. The TCN model favored a window of 40 (200~ms) and sequence length $T$=60 for classification. Stated differently, with a step-size of 25~ms, TCN accuracy was highest when its input sequence represented the preceding 1.675~s of EMG information. Importantly, the TCN contour shows that, for short sequences, performance is lower and more dependent on window length. For longer sequences, window length is less of a factor.}
		\label{params}    
	\end{figure}

	Temporal convolutional networks (TCN) are a class of sequential prediction models that are designed to learn hidden temporal dependencies within input sequences. The TCN model used herein consists of a single layer of convolutional filters followed by a time-distributed, fully-connected classification layer (Fig.~\ref{model}). Within the convolution layer, a collection of $M$=64 convolutional filters $\mathbf{W} \in\mathbb{R}^{d \times F}$, where $d$=25 is the duration of the filter and $F$=8 is the number of input features, are convolved along the temporal dimension of input sequence $\mathbf{X}\in\mathbb{R}^{F \times T}$, where $T$ is the number of 25~ms time-steps in the sequence, to produce temporal feature maps $\mathbf{E} \in\mathbb{R}^{M \times T}$, where
	\begin{align}	
		\mathbf{E}_i=f(\mathbf{W}_i * \mathbf{X}+ \mathbf{b}_i),\;\;\;\;\; i=1,\;2,\;\ldots,\;M.	
	\end{align} 
	Rectified linear unit (ReLU) activations~\cite{Hahnloser00} are applied to each element which are fed to a time-distributed, fully-connected layer for classification. Softmax activation~\cite{Luce59} is applied to produce $C$ class probabilities for current time $t$ 
	\begin{align}	
		\hat{\mathbf{y}}^{(t)} = \text{softmax}(\mathbf{UE}^{(t)}+ \mathbf{c})	
	\end{align}	
	where $\mathbf{U}\in\mathbb{R}^{C \times M}$ and $\mathbf{c}\in\mathbb{R}^{C\times 1}$ are the output weight matrix and bias, respectively. During preliminary testing, a subject (excluded from results) performed the experiment described in Sec. II-C. TCN and LSTM sequential models were trained from the first 3~min of data and tested on the last 3~min to determine optimal window sizes, input sequence lengths, and TCN dimensions. Performance contours for TCN and LSTM model parameters are shown in Fig.~\ref{params}.
			
\subsection{Assessing Prediction Stability}
	In addition to performance accuracy, we wish to quantify a sense of the \emph{stability} of a model, or how inclined the model is towards erratic \emph{class-switching} or misclassification during volitional class-to-class movement. Furthermore, a model may cleanly and appropriately switch between classes, but with slight delay or anticipation. The accuracy metric penalizes this generally benign behavior, often more-so than erratic behavior. 
	
	To complement performance accuracy, we define a stability metric to quantify a model's class-switching behavior relative to ground truth behavior. Given a vector $\mathbf{p}$ representing a series of $N\in\mathbb{Z}_{>1}$ predictions over time, we can count how many times that the prediction model switches its class output with 
	\begin{align}
		c_\mathbf{p} = \sum_{i=2}^{N}\big(1-\delta(p_i,p_{i-1}) \big)
	\end{align} 
	where $\delta(\cdot,\cdot)$ is the Kronecker delta function, or equivalence indicator. After computing $c_\mathbf{p}$ as well as $c_\mathbf{t}$ from ground truth label sequence $\mathbf{t}$, our prediction stability metric 
	\begin{align}
		S_\mathbf{p|t} = 1-\frac{|c_\mathbf{p}-c_\mathbf{t}|}{N-1} 
	\end{align} 
	quantifies how varied predictions are relative to ground truth. Ideally, $S_\mathbf{p|t}=1$, meaning a model's prediction output changes exactly as often as ground truth changes. Together, accuracy and our stability metric provide a fuller understanding of a model's behavior.  Our later comparison of TCN and LSTM highlights the potential for disparity between these metrics.  
	
\subsection{Experimental Methods}
	This study was conducted in accordance with protocols approved by the Johns Hopkins Medicine Institutional Review Boards. 9 able-bodied subjects (8 male, 1 female) participated in these experiments, ages: 24.1$\pm$3.2 years. Most subjects were inexperienced with EMG classification. 
	
\subsubsection{Data Acquisition, Prediction, and Visualization}
\begin{figure}[t!]
		\centering
		\includegraphics[width=.48\textwidth]{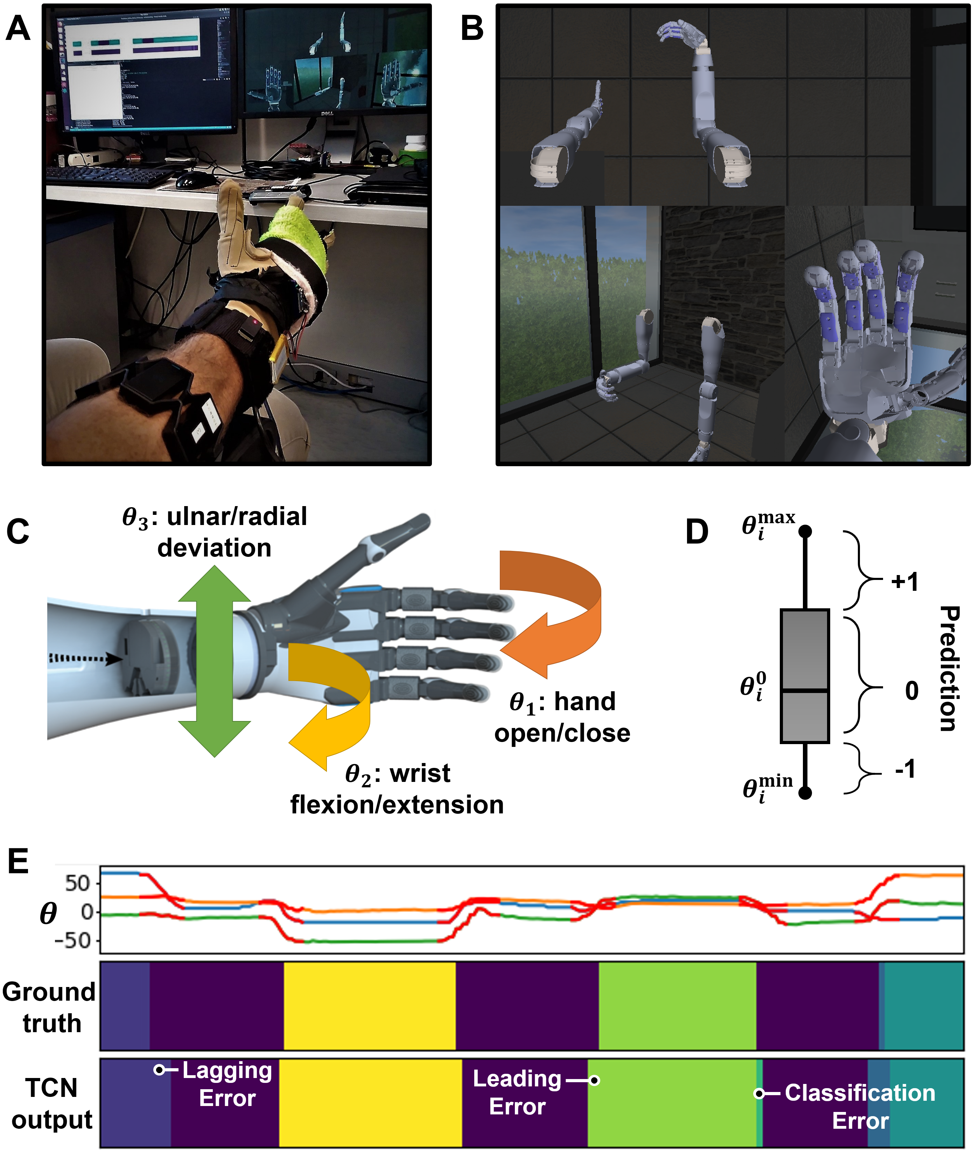}
		\caption{(A) Each subject was fitted with a Myo Armband, CyberGlove, and inertial sensors for EMG and hand/wrist positional recording. (B) The vMPL environment provided a real-time display of the subject's hand/wrist orientations. (C) EMG signals were used to predict movement classes along 3 hand/wrist degrees-of-freedom. (D) At each time-step, each DOF $i$ is converted from continuous joint position $\theta_i$ into a ternary class encoding (rest: 0, forward: +1, or reverse:-1) representing one of $27$ simultaneous 3-DOF movement classes. (E) Example sequence of 3-DOF joint angles $\mathbf{\theta}$, the corresponding conversion into ground truth class labels, and the class prediction output stream of TCN during this sequence. Three transient prediction problems are evident from this example: lagging, leading, and classification errors. The first two relate to specific timing of volitional movements, whereas the latter is an unintended movement class. The accuracy metric penalizes all three, whereas our defined stability metric only penalizes unintentional class-switching.}
		\label{exp}
	\end{figure}

	Eight channels of raw EMG sampled at 200~Hz were obtained from a Myo Armband (Thalmic Labs, Ontario, Canada) placed around the circumference of forearm muscle of greatest mass. Hand grasp position data were recorded with the Cyberglove II (CyberGlove Systems LLC, San Jose, CA). Wrist position data were recorded using two 9-axis MPU-9150 inertial sensors (InvenSense, San Jose, CA) (Fig.~\ref{exp}A). 
	
	For non-sequential models, every 25~ms we used a 200~ms sliding window to extract time-domain (TD5) features from the raw EMG signals: mean absolute value (MAV), waveform length, variance, slope sign change, and zero crossings~\cite{Hudgins93}. For sequential models TCN and LSTM, the optimal sliding window sizes were 200~ms and 175~ms, respectively. We observed that sequential models often performed best with only MAV features trained 35 epochs. In our results, we compared the TCN model (MAV only) with LSTM (MAV), as well as the following non-sequential prediction models (TD5): 
	\begin{table}[h!]\centering
			\begin{tabular}{ll}
				\midrule
				$k$-NN: & $k$-nearest neighbors, $k=3$\\
				SVM-RBF: & Support vector machine, gaussian radial basis\\ 	
				Forest: & Random forest\\
				ANN: & Artificial neural network, 3 layers x 5 nodes\\
				\midrule
		\end{tabular}
	\end{table}	

	\begin{figure*}[t!]
		\centering
		\includegraphics[width=.98\textwidth]{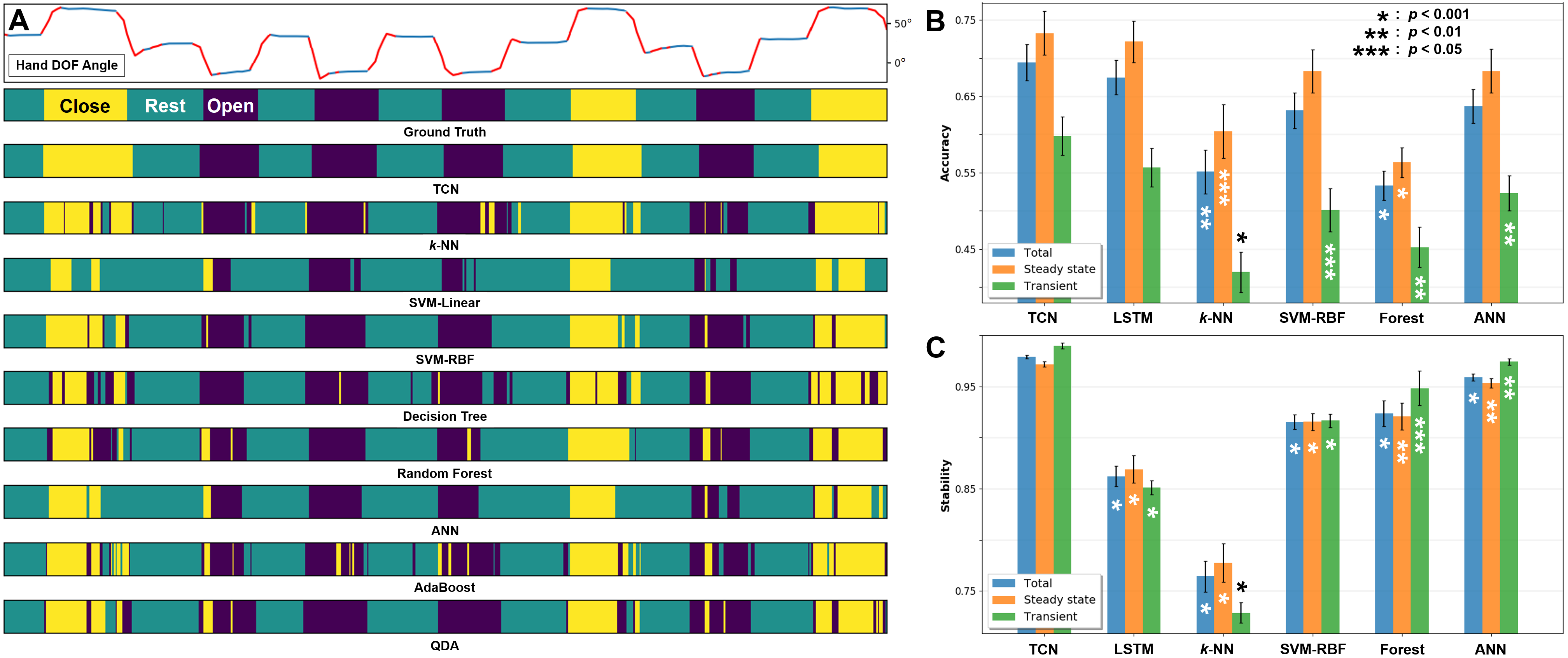}
		\caption{Comparative analysis of TCN with other classifiers. (A) Example prediction output behaviors of many popular classifiers when predicting 3 classes for a single DOF. Incorrect predictions often occur during interclass transitions. TCN demonstrated resilience in transient-states, but its class-switching was sometimes slightly anticipatory or delayed, informing our decision to devise a stability metric to complement the accuracy metric. (B) Prediction accuracy of 9 able-bodied subjects in our 3-DOF simultaneous experiment. Statistical significance thresholds are denoted. (C) Prediction stability, 9 able-bodied subjects. TCN and LSTM achieved higher accuracy with MAV features than non-sequential methods with TD5 features, though LSTM was less stable than TCN.}
		\label{offline}
	\end{figure*} 

   A user interface was developed for Python to control the virtual Modular Prosthetic Limb (vMPL) subcomponent of Johns Hopkins University Applied Physics Laboratory's MiniVIE system~\cite{vmpl13}\cite{Wester14} in order provide the subject with a real-time display of their current hand/wrist orientations (Fig.~\ref{exp}B). 

\subsubsection{3-DOF Simultaneous Protocol}
	After an initial 15~min practice session to familiarize the subject with movement classes and contraction consistency, subjects were asked to explore for 40~s their full range of motion in each of the 3 DOFs representing wrist and hand movements: rest, hand open/close, wrist flexion/extension, and radial/ulnar flexion (Fig.~\ref{exp}C). Outer boundary positions $\theta^{min}_i$, $\theta^{max}_i$, and comfortable interior rest position $\theta^0_i$ were determined for each DOF $i$ from this calibration period. Class thresholds along each DOF were set at 50\% of the distances from $\theta^0_i$ to $\theta^{min}_i$ and $\theta^{max}_i$. In other words, for the ground truth to switch from ``rest'' to ``hand close,'' the hand must be moved more than half the distance from rest to $\theta^{max}_1$. Thus, for every time step, each DOF records a ternary encoding (Fig.~\ref{exp}D) allowing for $27$ distinct simultaneous 3-DOF movement classes. We determined when the subject was in a transient-state by calculating where joint-velocity magnitude exceeded a threshold (shown as red in Fig.~\ref{exp}E). 
	
	Subjects were instructed to explore 3-DOF simultaneous movements held at consistent, repeatable contraction levels in freeform fashion (any order, combination, and duration $<$5~s) for 6~min while EMG and position data were recorded. In our analyses, the first 3~min of this EMG data were used for model training and the last 3~min for model testing. Since the boundaries of each $\theta_i$ were subject-determined, there was no need for movement-cue presentation nor a guarantee that the subject would attempt all 27 3-DOF combinations. 
	
\subsubsection{Experiment Analysis and Evaluation}
	All computations were performed with common Python 3.6.5 modules and the TemporalConvolutionalNetworks~\cite{TCN18} open-source package. Statistical $p$-values were computed from one-way analyses of variance (ANOVA) comparing TCN accuracy and stability with other models. Figure error-bars represent 1 standard error. 

\section{Results}
	Aggregated results from our 3-DOF simultaneous experiment, including $p$-values, are shown in Fig.~\ref{offline}. In general, TCN using only MAV features demonstrated significantly higher transient-state accuracy than non-sequential models using TD5 features: $p<0.001$ for $k$-NN; $p<0.01$ for Random forest and ANN; and $p<0.05$ for SVM-RBF (Fig.~\ref{offline}B). Furthermore, the stability of TCN was significantly higher than all other sequential and non-sequential models tested in both steady-states and transient-states (Fig.~\ref{offline}C). Importantly, the TCN and LSTM sequential models were similarly accurate, but LSTM was one of the least stable models. 
	
	For the most stable models (TCN, SVM, Forest, ANN), we observed that steady-state predictions were somewhat less stable on average (though consistently more accurate) than transient-states. These differences were not significant, but could indicate that some models become more unstable during pre-transition activation or post-transition recovery than during the physical transition itself. Examples of pre-transition instability can be seen in Figs. \ref{problems} and \ref{offline}A.
			

\section{Discussion and Conclusions}
	Instability in the prediction output stream is a well-known problem in EMG classification, particularly during transient interclass movement. Past attempts to mitigate instability such as majority filters and confidence-based rejection~\cite{Scheme13} focus primarily on post-processing the output of the classification model. In the specific case of majority filtration, the cost for improved stability is a prediction delay. Coupled with the EMG window length, this delay may be quite perceptible to the user. Confidence-based rejection is highly useful because it creates almost no time delay and can be appended to improve the stability of any model, including TCN. 
	
	To address inherent model stability, we hypothesized that sequential models, designed to utilize the temporal context of sequential input data, would significantly improve EMG classification compared with traditional non-sequential models (Fig.~\ref{offline}A). Notably, sequential models yielded better performance accuracy when provided with only the MAV feature for each channel, whereas all non-sequential models preferred the TD5 feature set (Fig.~\ref{offline}B). This indicates that the hidden temporal features learned from MAV sequences are equally or more valuable for EMG classification than non-sequential prediction of TD5 features. 
	
	TCN and LSTM perform similarly with respect to classification accuracy (Fig.~\ref{offline}B), but TCN provides significantly more stable output behavior compared to all models evaluated during both steady-states and transient-states (Fig.~\ref{offline}C). Therefore, though some loss in TCN accuracy is due to anticipation or delay in the timing of class-switching, its consistently stable behavior is very desirable for reliable control of upper-limb prostheses. 
	
	For natural prosthesis control, it is necessary to accurately predict during dynamic transient-states because our limbs are often in motion, between states, not merely switching discretely between fixed positions. The ability of TCN models to correctly predict during transient-state movements hints at very promising behavior when applied to multi-DOF regression, a research avenue we are currently exploring.
	
\addtolength{\textheight}{-0cm}	

\section*{Acknowledgment}
	We wish to acknowledge Johns Hopkins Applied Physics Laboratory (JHU/APL) for developing and making available the Virtual Integration Environment (VIE), which was developed under the Revolutionizing Prosthetics program.  This material is based upon work supported by the Defense Advanced Research Projects Agency (DARPA) under Contract No. N66001-10-C-4056.  Any opinions, findings and conclusions or recommendations expressed in this material are those of the author(s) and do not necessarily reflect the views of DARPA and JHU/APL. We thank the human subjects who participated in this study and our colleagues Dr. Brock Wester, Robert Armiger, Dr. Colin Lea, Tae Soo Kim, and Dr. Austin Reiter. 


	\bibliographystyle{ieeetr}

\end{document}